\newcommand{\brid}{\ensuremath{{\ensuremath{\hat \beta_{ridge}(\tau)     }}}}
\newcommand{\bridj}{\ensuremath{{\ensuremath{\hat \beta_{ridge,j}(\tau)     }}}}
\newcommand{\proj}{\ensuremath{\mathscr{P}_\tau}}
\newcommand{\projz}{\ensuremath{\mathscr{P}_0}}
\newcommand{\pen}{\ensuremath{\mbox{pen}}}

\newcommand{\pre}{\ensuremath{F}}
\newcommand{\pret}{\ensuremath{F}_\tau}

\newcommand{\R}{\ensuremath{{\mathbb{R}}}}
\newcommand{\X}{\ensuremath{{\mathbf{X}}}}

\newcommand{\bols}{\ensuremath{{\hat \beta^{ols}}}}

\newcommand{\bolsj}{\ensuremath{{\hat \beta_j^{ols}}}}

\documentclass[12pt]{amsart}
\usepackage{geometry}                
\RequirePackage{natbib}
\usepackage{mathrsfs}
\usepackage{graphicx}
\usepackage{amsthm}
\usepackage{amssymb}
\usepackage{epstopdf}

\usepackage{ulem}
\usepackage{setspace}
\newtheorem{definition}{Definition}
\newtheorem{theorem}{Theorem}

\newtheorem{lemma}{Lemma}

\title[Preconditioning for classical relationships]{A note relating ridge regression and OLS p-values to preconditioned sparse penalized regression}
\author{Karl Rohe\\UW-Madison Department of Statistics}
\thanks{For helpful comments, thanks to Derek Bean, Zoe Russek, Sara Fernandes-Taylor, Ming Yuan,  Jerry Freidman, and Garvesh Raskutti.  Rohe is supported by NSF grant DMS-1309998.}

\begin{document}
\maketitle
\begin{abstract}
When the design matrix has orthonormal columns, ``soft thresholding" the ordinary least squares (OLS) solution produces the Lasso solution \citep{tibshirani1996regression}.  If one uses the Puffer preconditioned Lasso \citep{jia2012preconditioning}, then this result generalizes from orthonormal designs to full rank designs (Theorem 1).  Theorem 2 refines the Puffer preconditioner to make the Lasso select the same model as removing  the elements of the OLS solution with the largest p-values.  
Using a generalized Puffer preconditioner, Theorem 3 relates ridge regression to the preconditioned Lasso; this result is for the high dimensional setting, $p>n$.
Where the standard Lasso is akin to forward selection \citep{efron2004least}, Theorems 1, 2, and 3 suggest that the preconditioned Lasso is more akin to backward elimination.  These results hold for sparse penalties beyond $\ell_1$;  for a broad class of sparse and non-convex techniques (e.g. SCAD and MC+), the results hold for all local minima.

\end{abstract}

\section{Introduction}

Preconditioning is a classical computational technique in numerical linear algebra that creates fast algorithms.  Several papers have recently proposed and studied the ``preconditioned" Lasso \citep{paul2008preconditioning, huang2011variable, rauhut2011sparse, jia2012preconditioning, qian2012pattern, wauthier2013comparative}. Instead of accelerating standard Lasso algorithms, preconditioning the Lasso creates a new statistical estimator that retains several properties of the Lasso while making the solution less sensitive to the correlation between the columns of the design matrix.  This paper demonstrates how penalized least squares estimators with various forms of preconditioning are equivalent to classical quantities in linear regression--the OLS estimator, OLS p-values, and ridge regression.

 The theorems below do not make any assumptions on the design matrix beyond full rank.  Nor do they assume a linear model $Y = \X \beta + \epsilon$, or assume some conditions on an error term $\epsilon$.  Instead of studying the statistical estimation properties of preconditioned penalized least squares problems, the following theorems study the preconditioned Lasso estimators as functions of data $(\X, Y)$ that return a vector $\hat \beta$.  The theorems compare these new functions to classical ``functions" like $ols$ and ridge regression.

\subsection{Preliminaries}
While the theorems below do not require the linear model, it is the linear model that motivates the estimators (i.e. functions) studied in this paper. The linear model is
\begin{equation}\label{linearModel}
Y = \X \beta + \epsilon,
\end{equation}
where $Y \in R^n$ and $\X \in R^{n \times p}$ are observed, and $\epsilon \in R^n$ is random noise satisfying $E(\epsilon) = 0$ and $E(\epsilon \epsilon') = \sigma^2 I_p$.  The goal is to estimate $\beta \in R^p$ with $\X$ and $Y$.   

Throughout the paper, we will assume that $\X$ is full rank (when $n<p$, it is full row rank). Define $\|x\|_q = (\sum_i x_i^q)^{1/q}$.
For $n>p$, define 
 \[\hat \beta^{ols} =  \arg \min_{b\in R^p} \|  Y -   \X b\|_2^2 = (\X'\X)^{-1} \X' Y \in R^p\]
 as the standard OLS estimator.
\begin{definition}
Define the function $Lasso_\lambda$,
\[Lasso_\lambda(  \X,   Y) = \arg \min_{b\in R^p} \|  Y -   \X b\|_2^2 + \lambda \|b\|_1.\]
\end{definition}

For $a\in R$, define $\textrm{sign}(a)$ as $-1, 0,$ or $1$, depending on whether $a$ is negative, zero, or positive.  Define $(a)^+$ as equal to $a$ if $a$ is nonnegative and equal to zero if $a$ is negative.  The soft-thresholding function is defined as 
\[t_\lambda(x) = sign(x) (|x| - \lambda)^+.\]
To apply $t_\lambda$ to a vector $x$, apply it element-wise, $[t_\lambda(x)]_j = t_\lambda(x_j)$. 

\begin{lemma}\label{classicalRelation} (Equation 3 in \citep{tibshirani1996regression})
If the design matrix $\X$ is orthonormal, 
\begin{equation} \label{classical}
Lasso_\lambda(\X, Y) = t_\lambda(\bols).
\end{equation}
\end{lemma}

%


\section{Preconditioning the Lasso}
Sections 2 and 3 study the low dimensional setting $n > p$.  For these sections, let   $\X = UDV'$ be the ``skinny" SVD; $U \in R^{n \times p}$ and $V\in R^{p \times p}$ have orthonormal columns and $D$ is a diagonal matrix.  The Puffer transform is defined as $\pre = U D^{-1} U'$ \citep{jia2012preconditioning}.  After preconditioning, Equation \eqref{linearModel} becomes
\begin{equation} \label{precondModel}
\pre Y = (\pre \X) \beta + \pre \epsilon.
\end{equation}
 While $(\pre \X) =U V'$ is an orthonormal matrix, it is not orthogonalized by rotating the columns as in a QR decomposition; this would correspond to \textit{right} multiplying $\X$ by some matrix.  Rotating the columns would create a new basis and make the Lasso penalize in the incorrect basis.  By \textit{left} multiplying, each row of $\pre \X$ is a linear combination of the rows in $\X$. 
Importantly, the regression estimators that use $(\pre \X, \pre Y)$ instead of $(\X, Y)$ still estimate the same vector $\beta$ and the Lasso penalizes in the correct basis; the original regression model in Equation \eqref{linearModel} contains the exact same $\beta$ as Equation \eqref{precondModel}. Define 
\begin{eqnarray*}
\textit{Puffer}(\X, Y) &=& (\pre \X , \pre Y) \in \R^{n \times p} \times \R^n \ \mbox{ and } \\ 
Lasso_\lambda(\textit{Puffer}(\X, Y)) &=& Lasso_\lambda(\pre \X, \pre Y). 
\end{eqnarray*}
\cite{jia2012preconditioning} showed that if the smallest singular value of $\X$ is bounded from below, then with $p$ fixed and $n\rightarrow \infty$, the preconditioned Lasso, $Lasso_\lambda(\textit{Puffer}(\X, Y))$, is sign consistent.  Importantly, the Puffer preconditioned Lasso does not require the Irrepresentable Condition from \cite{zhao2006model}.  The next theorem shows how this estimator relates to the classical estimator $\hat \beta^{ols} \in R^p$, computed on the full model.  This  extends the relationship in Lemma \ref{classicalRelation} from the case where $\X$ is orthonormal, to the case where $\X$ is full rank.




\begin{theorem}\label{theorem1}
If $\X$ is full rank and $n>p$, then
\begin{equation}\label{threshold}
Lasso_\lambda(\textit{Puffer}(\X,  Y)) = t_\lambda(\hat \beta^{ols} ).
\end{equation}
\end{theorem}
  In general, the relationship in Equation \eqref{threshold} does not hold without \textit{Puffer}.
All proofs are contained in the Appendix.
%



\section{Correcting for the heterogeneous variability in $\hat \beta^{ols}$}

Under the linear model, 
\begin{equation}\label{olsSE}
cov(\bols) = \sigma^2 (\X'\X)^{-1}.
\end{equation}
Define 
\[\Sigma^{ols} = cov(\bols). \ \mbox{ Then, } variance(\bolsj) = \Sigma_{jj}^{ols}.\]
If the diagonal elements of $\Sigma^{ols}$ are not all equal, then some elements of $\bols$ will have greater variability than others.  Classical confidence intervals and p-values account for this uncertainty.  However, the preconditioned Lasso estimator above (i.e. $t_\lambda( \bols) $) does not account for this known heteroskedasticity  of $\bols$ by applying a stronger penalty to terms with larger variance.

Classical versions of model selection test the following null hypotheses in various ways:
\[H_{0,j}: \beta_j = 0,  \ \mbox{ for } j \in 1, \dots, p.\]
Under the linear model \eqref{linearModel} and $H_{0,j}$, 
\begin{equation}\label{teststat}
Z_j = \sqrt{n} \frac{\hat \beta_j^{ols}}{\sqrt{\sigma^2 (\X'\X)_{jj}^{-1}}} \stackrel{d}{\Rightarrow}N(0,1) \ \mbox{ as $n \rightarrow \infty$}.
\end{equation}
The classical ``marginal" p-values are defined as
\begin{equation}\label{pvalue}
p_j = 2(1 - \Phi(|Z_j|)), \mbox{ for $j = 1, \dots, p$}
\end{equation}
where $\Phi$ is the cdf of the standard normal distribution. 

\subsection{Scaling matters}

Standard Lasso packages normalize the columns of $\X$ so that they all have equal $\ell_2$ length. This ensures that each element of the Lasso estimator is equally penalized.  However, it does not ensure that the elements of the estimator are equally variable. This section  \textit{right} preconditions $\X$ with a diagonal matrix $N$, making the column lengths heterogeneous.   The column lengths are chosen so that the penalty strengths align with the heteroskedasticity of $\bols$, ensuring equal variability among the elements of the estimator $\hat \beta$.  

After left and right preconditioning with $\pre$ and $N$ respectively, the regression equation becomes
\[\pre Y = (\pre \X N)(N^{-1}\beta) + \pre \epsilon.\]
Define $\nu \in R^p$ as the diagonal elements of $(\X' \X)^{-1}$ and let $N \in R^{ p \times p}$ be a diagonal matrix with 
$N_{jj} = \sqrt{\nu_j}. $
Because $\X$ is assumed full column rank and $N_{jj}$ is proportional to the standard error of $\bolsj$, $N_{jj}$ exists and is strictly positive.  This ensures that $\textrm{sign}(N^{-1}\beta) = \textrm{sign}(\beta).$ So, inferences in the transformed space carry over to the original model.

Take the SVD of $\X N = U_N D_N V_N'$. Then, define $\pre_N = U_N D_N^{-1} U_N'$ and 
\[\textit{Puffer}_N(\X, Y) = (\pre_N \X N, \pre_N Y) \in \R^{n \times p} \times \R^n.\]
Theorem \ref{pvalueLasso} shows that 
if $\lambda = 1.96 \sigma/\sqrt{n}$, then $Lasso_\lambda(\textit{Puffer}_N(\X, Y))$ selects the same variables as fitting the standard OLS estimator and removing any variables with p-value greater than $.05$.

\begin{theorem} \label{pvalueLasso}
If $\X$ is full rank and $n>p$, denote
\[\hat \beta^N(\lambda) = Lasso_\lambda(\textit{Puffer}_N(\X, Y)).\]
Let $Z_j$ be the classical test statistic defined in Equation \eqref{teststat}, let $p_j$ be the classical p-value defined in Equation \eqref{pvalue}, and let $\Phi$ represent the cdf of the standard normal distribution.  
\[\hat \beta_j^N(\lambda) \ne 0 \ \  \ \Leftrightarrow  \ \ \ |Z_j| > \lambda \sqrt{n} / \sigma \ \ \ \Leftrightarrow \ \ \ p_j \le 2\left(1-\Phi(\lambda \sqrt{n} /\sigma)\right).\]

\end{theorem}

Importantly, this is an algebraic equivalence between $Lasso_\lambda(\textit{Puffer}_N(\X, Y))$ and the classical OLS p-values in Equation \eqref{pvalue}.  This requires only a full rank design matrix $\X$ and does not make any assumption on the error distribution or the elements of $\beta$.  Theorem 2 asserts nothing about the statistical reliability of these p-values.  If one wishes to use the p-value for statistical inference, then Theorem \ref{pvalueLasso} needs additional assumptions on the error term, $\epsilon$.

%
%

%

%
%

\subsection{Generalizing to other methods}

For simplicity, Theorems 1 and 2 are stated for the Lasso.  However, both theorems hold more generally.
For $\lambda \ge 0$ and penalty function $\pen: \R \rightarrow \R_+$, define the function $sparse(\X,Y,\lambda, \pen)$ as 
\begin{equation}
sparse(\X,Y,\lambda, \pen) = \arg \min_b \frac{1}{2}\|Y - \X b\|_2^2 + \lambda \sum_j \pen(b_j).
\end{equation}
Whenever $\pen$ has a thresholding function $\tilde t_\lambda$ that satisfies a version of Lemma \ref{classicalRelation} with orthonormal designs, i.e.
\[sparse(\X, Y,\lambda, \pen) = \tilde t_\lambda(\bols) \ \mbox{ for orthonormal } \X,\]
then modified versions of Theorems 1 and 2 also apply for this penalty:
\begin{eqnarray*}
&sparse(\textit{Puffer}(\X, Y),\lambda, \pen) &= \tilde t_\lambda(\bols)  \\
&\left[ sparse(\textit{Puffer}_N(\X, Y),\lambda, \pen)\right]_j &= \tilde t_\lambda(  \sigma Z_j / \sqrt{n} ).
\end{eqnarray*}
For example, $\pen$ could be an $\ell_q$ penalty for $q \in [0,1]$, the elastic net penalty, or a concave penalty such as SCAD or MC+ \citep{fan2001variable, zhang2010nearly}.  After preconditioning, a broad class of penalties  select the same sequence of models as the Lasso.

\section{Relating to ridge regression}

For $p > n$, take the ``skinny" SVD of 
$\X = UDV'$, where $U \in R^{n \times n}, V \in R^{p \times n}, D \in R^{n \times n}$.  Consider a generalized Puffer transformation
\[\pret = U(D^2 + \tau I)^{-1/2}U' \ \mbox{ and } \textit{Puffer}_\tau(\X, Y) = (\pret \X, \pret Y).\]
Theorem \ref{mainresult} relates $sparse(\textit{Puffer}_\tau(\X, Y), \lambda, \pen)$  to  the ridge estimator \citep{hoerl1970ridge}.  For $\tau > 0$, the ridge estimator is 
\begin{eqnarray}
\brid &=& \arg \min_b \|Y - \X b\|_2^2 + \tau \|b\|_2^2 \label{ridge} \\
 &=& (\X'\X + \tau I)^{-1} \X' Y. \nonumber
\end{eqnarray}
Define $\hat \beta_{ridge}(0)$ as the Moore-Penrose estimator 
\begin{eqnarray*}
\hat \beta_{ridge}(0) &=&\arg \min_{b: \X b = Y}  \|b\|_2^2 \\
 &=& (\X'\X)^{+} \X' Y,
\end{eqnarray*}
where $(\X'\X)^{+}$ is the Moore-Penrose pseudoinverse, $VD^{-2}V'$.\footnote{In what follows, replace any matrix inversion with the Moore-Penrose pseudoinverse if $\tau =0$ and the true inverse does not exist.}

Define $\proj: \R^p \rightarrow \R^p$ as
\begin{eqnarray}
\proj(v) &=& \X'(\X\X' + \tau I)^{-1} \X v. \label{proj} 
\end{eqnarray}
For $\tau = 0$, $\projz$ projects onto the row space of $\X$.  Under the linear model (Equation \ref{linearModel}), $\brid$ is an unbiased estimator of $\proj(\beta)$.

The following theorem assumes that $sparse$ uses a penalty function satisfying the following assumptions.
\begin{definition} \label{sparsepen}
Define a function $\pen$ as a \textbf{regular sparse penalty} if it is 
\begin{enumerate}
\item non-differentiable at zero and differentiable everywhere else;
\item symmetric, $\pen(a) = \pen(-a)$; 
\item monotonically increasing away from zero, $\pen(a) \ge \pen(b)$ if $|a| > |b|$; 
\item Lasso derivative in the neighborhood of zero,
\[\lim_{x \rightarrow 0} |\pen'(x)| = 1,\] 
where $\pen'$ is the first derivative of $\pen$.
\end{enumerate}
\end{definition}
This includes the Lasso, elastic net, SCAD, and MC+.  However, $\ell_q$ penalties fail condition (4) when $q < 1$.  Such penalties have an unbounded derivative in the neighborhood of zero which creates discontinuities in the solution path;  previous research has also excluded such penalties (e.g. \cite{zhang2012general, loh2013regularized}).

Theorem \ref{mainresult} says that if $\pen$ is a regular sparse penalty, then any local minimum for the objective function in $sparse(\textit{Puffer}_\tau(\X, Y), \lambda, \pen )$, transformed by $\proj$, is close to the ridge estimator $\brid$.  Moreover, $\lambda$ controls the distance between these estimators.  

\begin{theorem} \label{mainresult}
Let $\pen$ be a regular sparse penalty (Definition \ref{sparsepen}) and let $p\ge n$.  Let 
\[\hat \beta = sparse(\textit{Puffer}_\tau(\X, Y), \lambda, \pen ).\] 
If $\hat \beta_j \ne 0$, then
\[\bridj - \proj(\hat \beta)_j = \lambda \pen'(\hat \beta_j),\]
where $\pen'$ is the derivative of $\pen$.  If $\hat \beta_j = 0$, then 
\[|\bridj - \proj(\hat \beta)_j| \le \lambda.\]
Moreover, these results still hold if $\hat \beta$ is any local minimizer of the objective function for $sparse(\textit{Puffer}_\tau(\X, Y), \lambda, \pen )$.
\end{theorem}

Importantly, $sparse$ has been computed with the preconditioned data, $\textit{Puffer}_\tau(\X, Y)$, while $\brid$ is the traditional estimator computed with the original data.  When $\lambda$ is small, these two estimators are aligned in the row space of $\X$.  Determining the statistically appropriate scale of $\lambda$ requires some care because after preconditioning, the scale of the problem changes; $\|\X\|_F^2 = O(np)$, but $\|\pret \X\|_F^2 = O(n)$.  A forthcoming revision to \cite{jia2012preconditioning},  shows that $sparse(\textit{Puffer}_0(\X, Y), \lambda, \| \cdot \|_1)$ is sign consistent when $\min_j \beta_j$ is larger than $\log n / \sqrt{n}$ and 
\[\lambda = O\left(\left(\frac{\log n \log p}{p} \right)^{1/2}\right).\]
In the high dimensional setting, this $\lambda$ is clearly converging to zero.

If Theorem 3 held without $\proj$, then it would say that $Lasso_\lambda(\textit{Puffer}_\tau(\X, Y)) = t_\lambda(\brid)$, which would make a clear analogy to backward elimination.  However, the inclusion of $\proj$ stains the analogy to backward elimination. In fact, it is not even true that $\proj\left(Lasso_\lambda(\textit{Puffer}_\tau(\X, Y))\right) = t_\lambda(\brid)$; the left side is in the row space of $\X$, but the right side is not. That said,
typical algorithms to compute the Lasso  solution path (e.g. \cite{efron2004least}) start at $\lambda =\infty$ with $Lasso_{\lambda= \infty}(\X, Y) = 0$ and decrease $\lambda$, increasing the number of terms in the model.  This resembles forward selection.
The results of Theorem 3 are ``backwards" in the sense that for $\lambda \rightarrow 0$, $\proj(Lasso_\lambda(\textit{Puffer}_\tau(\X, Y))) \rightarrow \brid$.

Suppose $\hat \beta^{(1)}$ and $\hat \beta^{(2)}$ are both local minima of $sparse(\textit{Puffer}_0(\X, Y),\lambda, \pen)$ for a regular sparse penalty $\pen(x)$ that is concave on $\{x:x>0\}$.  By concavity and the definition of regular sparse penalty, it follows that $|\pen'(x)| \le 1$ for $x\ne0$.  So, the triangle inequality around $\hat \beta_{ridge,j}(0)$ yields
\begin{equation}\label{closeMins}
\projz( \hat \beta^{(1)}_j  - \hat \beta^{(2)}_j ) \le 2 \lambda.
\end{equation}
So after preconditioning, local minima are exceedingly similar in the row space of $\X$.  Moreover, even if $\hat \beta^{(1)}$ and $\hat \beta^{(2)}$ are (local) minima from different penalty functions, then Equation \ref{closeMins} holds so long as (i) both $\pen$ functions are regular and concave and (ii) both are computed with the same tuning parameter $\lambda$.\footnote{If they have different tuning parameters $\lambda_1$ and $\lambda_2$, then the bound $2 \lambda$ is replaced by $\lambda_1 + \lambda_2$.}  In general, $\beta$ is not identifiable when $p>n$; only the projection of $\beta$ into the row space of $\X$ is identifiable \citep{shao2012estimation}.  This suggests that, after preconditioning, it is difficult statistically distinguish the difference between local minima or the difference between penalty functions.

\section{Discussion}

Several previous papers have studied different preconditioners for the Lasso \citep{paul2008preconditioning, huang2011variable, rauhut2011sparse, jia2012preconditioning, qian2012pattern, wauthier2013comparative}.  
This paper connects two types of preconditioning techniques to the classical OLS solution.  When performing model selection in the classical setting of $n>>p$, the marginal p-values from OLS are typically considered more informative than the absolute sizes of the elements in $\hat \beta^{ols}$; the p-values account for the potentially heterogeneous standard errors across $\hat \beta^{ols}$.  This suggests that $Lasso_\lambda(\textit{Puffer}_N(\X, Y))$ should be preferred to $Lasso_\lambda(\textit{Puffer}(\X, Y))$.  However, classical intuitions also suggest that two variables might have statistically insignificant p-values because these variables are correlated with each other(see Section 10.1 in \cite{weisberg2014applied}).  As such, a backward procedure should have multiple steps.  On each step, remove the variable with the largest p-value and refit the OLS with the remaining predictors.  Neither of the preconditioning techniques above  creates a Lasso solution path that is equivalent to this multi-step approach.  A preconditioner that is designed to match this path must be a function of $Y$ and $\lambda$, thus becoming much more complicated.

 The Lasso (without preconditioning) is akin to forward selection \citep{efron2004least}.  Classical methods of forward selection are not model selection consistent, unless the columns of the design matrix are only  weakly correlated \citep{tropp2007signal}.  Similarly, the Lasso is not sign consistent unless the design matrix satisfies the Irrepresentable Condition \citep{zhao2006model}.  After preconditioning with the (generalized) Puffer transformation, Theorems 1, 2, and 3 show that  the Lasso is akin to backward elimination.   This preconditioner makes the design orthogonal when $n>p$, trivially satisfying all consistency conditions (e.g. Irrepresentable Condition, RIP, etc.).  As such, Theorem 1 implies that a one step backward elimination is also sign consistent.  
 
This suggests that backward procedures are sign consistent when  forward procedures are not.
 However, backward procedures are not a panacea. They will become unstable whenever the OLS p-values are incorrect (e.g. when $p$ is large compared to $n$  and the CLT does not hold for the test statistic $Z_j$ in Equation \eqref{teststat}).



\bibliographystyle{plainnat}
\bibliography{references}

\appendix
\section{}
\textbf{Proof for Theorem 1}
\begin{proof}
Define the function $ols$,
 \[ols(\tilde \X,\tilde Y) = (\tilde \X'\tilde \X)^{-1}\tilde \X'\tilde Y.\]
 $\pre \X$ has orthonormal columns. So, by Lemma \ref{classicalRelation},
\[Lasso_\lambda(\pre \X, \pre Y) = t_\lambda( ols(\pre \X, \pre Y) ).\]  
Again using the fact that $\pre \X$ has orthonormal columns, $ ols(\pre \X, \pre Y) = \hat \beta^{ols}$;
 \begin{eqnarray*}
 ols(\pre \X, \pre Y) &=& ((\pre \X)' \pre \X)^{-1} (\pre \X)' \pre Y\\
&=&  (\pre \X)' \pre Y \\ 
& =& VD^{-1}U'Y \\
&=& (VD^2V')^{-1}VDU'Y  = (\X'\X)^{-1}\X'Y = \hat \beta^{ols}.
 \end{eqnarray*}

\end{proof}

\textbf{Proof for Theorem 2}
\begin{proof}
Define $\X_N = \X N$; its Puffer transformation is $\pre_N$. 
Using Theorem \ref{theorem1}, 
\begin{eqnarray*}
\hat \beta^N(\lambda) = Lasso_\lambda(\pre_N \X N, \pre_N Y) &=& Lasso_\lambda(\pre_N \X_N, \pre_N Y) \\
&=&  t_\lambda(ols( \X_N , Y)) = t_\lambda(ols( \X N , Y)). 
\end{eqnarray*}
Then,
\[ols( \X N , Y) = \left((\X N)' (\X N)\right)^{-1} (\X N)' Y= N^{-1}(\X'\X)^{-1}N^{-1} N X' Y = N^{-1} \hat \beta^{ols}\]
and the $j$th element of this is 
\[[ols( \X N , Y)]_j  
= \frac{\hat \beta_j^{ols}}{\sqrt{(\X' \X)_{jj}^{-1}}} =  \sigma Z_j / \sqrt{n} ,\]
where $Z_j$ is the test statistic for $H_{0,j}$ defined in Equation \eqref{teststat}. 

Putting this together, $[\hat \beta^N(\lambda)]_j = t_\lambda(  \sigma Z_j / \sqrt{n} )$. So, 
\[\hat \beta_j^N(\lambda) \ne 0  \ \Leftrightarrow  \ |Z_j| > \lambda \sqrt{n} / \sigma \ \Leftrightarrow \ p_j \le 2\left(1-\Phi(\lambda \sqrt{n} /\sigma)\right).\]
\end{proof}
 
\textbf{The proof for Theorem 3}
relies on the following lemma.
\begin{lemma} \label{ridgelemma}
If $p\ge n$, then for any vector $v \in \R^p$,
\begin{eqnarray*}
\proj(v)  &=& (\pret \X)' \pret \X v\\
\brid &=& (\pret \X)' \pret Y,
\end{eqnarray*}
where 
$\proj$ is defined in Equation \eqref{proj}, and $\brid$ is defined in Equation \eqref{ridge}.
\end{lemma}
A proof of Lemma \ref{ridgelemma} follows this proof of Theorem \ref{mainresult}.
\begin{proof}
If $\hat \beta$ is a local minimizer of 
\[ \frac{1}{2}\|\pret Y - \pret \X b\|_2^2 + \lambda \sum_j \pen(b_j),\]
then,  
\begin{equation} \label{kkt}
( \pret \X)'\pret Y - (\pret \X)' \pret \X \hat \beta - \lambda \hspace{.05cm} \partial \hspace{.05cm} \pen(\hat \beta) = 0,
\end{equation}
where $\pen(\hat \beta) \in \R^p$ is defined as $[\pen(\hat \beta)]_j = \pen(\hat \beta_j)$ and $\partial \hspace{.05cm} \pen(\hat \beta)$ is a generalized subgradient of $\pen(\hat \beta)$ \citep{clarke1990optimization}.  By the assumption that $\pen$ is a regular sparse penalty, if $x \ne 0$, then $\partial \hspace{.05cm} \pen(x) = \pen'(x)$ and $\partial \hspace{.05cm} \pen(0) \in [-1,1]$.  

Substituting the results from Lemma \ref{ridgelemma} into Equation \ref{kkt} gives the result.  

\end{proof}

The following proves Lemma \ref{ridgelemma}.
\begin{proof}
When $p\ge n$, the matrix $U \in \R^{n \times n}$ is orthonormal.  So,
\[\pret' \pret = U ( D^2 + \tau I)^{-1} U' = (UD^2U' + \tau UU')^{-1} = (\X\X' + \tau I)^{-1}.\]
Using this,
\begin{eqnarray*}
\proj(v)  &=& \X'(\X\X' + \tau I)^{-1} \X v\\
&=& \X' \pret' \pret \X v\\
&=& (\pret \X)' \pret \X v.
\end{eqnarray*}

Let $\X = U \tilde D \tilde V'$   be the ``full" SVD with $\tilde V \in \R^{p \times p}$ and $\tilde D \in \R^{n \times p}$.  
Notice that $U$ is unchanged. In the following calculations, the identity matrix $I$ takes a subscript denoting its dimension, $I_d \in \R^{d \times d}$.

\begin{eqnarray*}
(\pret \X)' \pret Y 
&=& VDU' U ( D^2 + \tau I_n)^{-1} U' Y\\
&=& V( D^2 + \tau I_n)^{-1} D U' Y\\
&=& \tilde V( \tilde D' \tilde D + \tau I_p)^{-1} \tilde D' U' Y\\
&=& \tilde V( \tilde D' \tilde D + \tau I_p)^{-1}\tilde V' \tilde V \tilde D' U' Y\\
&=& ( \tilde V \tilde D' \tilde D \tilde V' + \tau \tilde V\tilde V')^{-1} \tilde V \tilde D' U' Y\\
&=& ( \X'\X + \tau I_p)^{-1} \X' Y\\
&=& \brid.
\end{eqnarray*}


\end{proof}

\end{document}